\pdfoutput=1

\documentclass[11pt]{article}

\usepackage{emnlp2021}

\usepackage{times}
\usepackage{latexsym}

\usepackage[T1]{fontenc}

\usepackage[utf8]{inputenc}

\usepackage{microtype}

\usepackage{lastpage}
\usepackage{fancyhdr}
\fancypagestyle{plain}{%
  \fancyhf{}%
  \fancyfoot[C]{\thepage\ of \pageref{LastPage}}}

\title{Multilingual Search with Subword TF-IDF}

\author{Artit Wangperawong \\
  \texttt{artitw@gmail.com} \\}

\begin{document}
\maketitle
\thispagestyle{fancy}
\begin{abstract}
Multilingual search can be achieved with subword tokenization. The accuracy of traditional TF-IDF approaches depend on manually curated tokenization, stop words and stemming rules, whereas subword TF-IDF (\textbf{STF-IDF}) can offer higher accuracy without such heuristics. Moreover, multilingual support can be incorporated inherently as part of the subword tokenization model training. XQuAD evaluation demonstrates the advantages of STF-IDF: superior information retrieval accuracy of 85.4\% for English and over 80\% for 10 other languages without any heuristics-based preprocessing. The software to reproduce these results are open-sourced as a part of \href{https://github.com/artitw/text2text}{Text2Text}.
\end{abstract}

\section{Introduction}
Search and information retrieval systems that require bespoke setup can become expensive to modify and maintain. As both the collection of documents and the consumer behavior of such services evolve over time, working heuristics can become outdated due to changing data distributions and application requirements. These are also known as \textit{data drift} \citep{quinonero2008dataset} and \textit{concept drift} \citep{widmer1996learning}, respectively.

These problems are exacerbated for non-English languages, since tokenization rules can be significantly different, and there are less publicly available resources. Moreover, strategies developed for English are not universally transferable to other languages. For example, there are no whitespace delimiters for splitting sentences into words for Thai. Here we present a machine learning approach that obviates the need for heuristics-based text preprocessing. We propose \textbf{STF-IDF}: subword tokenization used in conjunction with TF-IDF on multilingual corpora.

\section{Evaluation Approach}
We focus our evaluation on the Cross-lingual Question Answering Dataset (XQuAD) \citep{Artetxe:etal:2019}, which consists of Wikipedia articles divided into 240 paragraphs. Each paragraph is associated with multiple question-answer pairs, totalling 1190 across all paragraphs. The dataset is paralleled across English, Spanish, German, Greek, Russian, Turkish, Arabic, Vietnamese, Thai, Chinese, Hindi, and Romanian. 

Each question is used as a query on all the paragraphs. Correctness is then determined by whether the model under evaluation returns the matching paragraph for the given question query, since the paragraph is supposed to contain the answer to the question. For each language, the accuracy is then the total number of correct matches divided by the total number of question queries for that language (1190).

\section{TF-IDF}
The TF-IDF vector space model -- representing queries and documents as vectors -- is an established ranked retrieval method for search and mining applications \citep{jones1972statistical}. The cosine similarity score between a query vector and a document vector can be used to determine how well the pair matches in comparison to other pairs. In practice, the vectors are derived from language-specific, heuristics-based processing. For English, word tokenization followed by stop word removal and stemming is a common pattern.

\subsection{Word Splitting}
Word tokenization involves splitting a sentence on whitespace delimiters. Identifying whitespace delimiters is also commonly based on heuristics. Depending on the application, spaces, tabs, carriage returns, dashes, slashes, special characters, and punctuation characters can be considered whitespace. Upon splitting sentences into word tokens, each token is assigned a common id or a position across the TF-IDF vectors for the documents. The frequency of each token in a document $TF$ is then multiplied by the inverse document frequency of the token across all documents $IDF$. For XQuAD's English passage retrieval, word tokenization alone can achieve 84.2\% accuracy as shown in Table~\ref{tab:tfidf}.

\subsection{Stop Word Removal}
Stop word removal involves filtering out words which are not considered useful for providing signal in the scoring process. Stop words are typically agglomerated into a list which the maintainer can add to or remove from as appropriate for the application. Although there is not a universally applicable stop words list, in practice it is not uncommon to use a starter list containing the most frequently used stop words. 

For XQuAD's English passage retrieval, word tokenization followed by stop word removal results in a lower accuracy (83.9\%) than word tokenization alone as shown in Table~\ref{tab:tfidf}. This could be due to reliance on a starter stop word list without any customization for the particular dataset. In any case, we demonstrate in the following section that even removing default stop words can be advantageous when used in conjunction with stemming.

\subsection{Stemming}
Stemming is the process of normalizing inflected or derived words to their word stem, which is not necessarily identical to the morphological root of the word. The purpose is to ensure that words with the same meaning are treated as such. The Porter Stemmer is a popular algorithm for stemming English words \citep{porter1980algorithm}. For XQuAD's English passage retrieval, word tokenization followed by stemming results in a higher accuracy (84.9\%) than word tokenization alone as shown in Table~\ref{tab:tfidf}. 

\begin{table}
\centering
\begin{tabular}{lc}
\hline
\textbf{Tokenization} & \textbf{Accuracy (\%)}\\
\hline
{word} & {84.2} \\
{word > stop} & {83.9} \\
{word > stem} & {84.9} \\ 
{word > stop > stem} & {85.2} \\ 
\hline
\end{tabular}
\caption{Accuracy comparison of TF-IDF tokenization strategies on XQuAD's English passage retrieval.}
\label{tab:tfidf}
\end{table}

\begin{table}
\centering
\begin{tabular}{lc}
\hline
\textbf{Tokenization} & \textbf{Accuracy (\%)}\\
\hline
{subword} & {85.4} \\
{word > stop > subword} & {84.2} \\
{word > stem > subword} & {85.4} \\ 
{word > stop > stem > subword} & {84.5} \\ 
\hline
\end{tabular}
\caption{Accuracy comparison of STF-IDF tokenization strategies on XQuAD's English passage retrieval.}
\label{tab:stfidf}
\end{table}

\begin{table}
\centering
\begin{tabular}{lc}
\hline
\textbf{Language} & \textbf{Accuracy (\%)}\\
\hline
{English (en)} & {85.4} \\
{Spanish (es)} & {85.8} \\
{German (de)} & {84.9} \\ 
{Greek (el)} & {81.3} \\ 
{Russian (ru)} & {82.9} \\
{Turkish (tr)} & {80.1}  \\ 
{Arabic (ar)} & {77.1}  \\
{Vietnamese (vi)} & {84.5} \\ 
{Thai (th)} & {83.5} \\
{Chinese (zh)} & {82.4}  \\ 
{Hindi (hi)} & {80.9}  \\
{Romanian (ro)} & {85.0}  \\
\hline
\end{tabular}
\caption{Accuracy comparison of STF-IDF on XQuAD passage retrieval of various languages.}
\label{tab:mstfidf}
\end{table}

By both removing stop words and stemming, we are able to achieve an even higher accuracy of 85.2\%. This enhancement could be attributed to stop words being useful only when unstemmed, but stemming all words being ultimately superior. For these empirical reasons, it is common practice to perform all three steps: word splitting, stop word removal, and word stemming.

\section{STF-IDF}

\subsection{Subword Tokenization}
Subword tokenization can be advantageous for multiple reasons \citep{sennrich2015neural}, including support for out-of-vocabulary (OOV) words. When an incoming query contains a word that does not exist in a word-based dictionary in traditional TF-IDF trained on a corpus of documents, the word is ignored. STF-IDF can utilize OOV words by breaking it down into the subword vocabulary.

Subword tokenization can account for internal structure of words that relate to
\begin{enumerate}
  \item names via character copying or transliteration,
  \item compounds via compositional translation, and
  \item cognates and loanwords via phonological and morphological transformations.
\end{enumerate}

Finally, subword tokenization can offer a single vocabulary for all languages. This allows not only multilingual support out of the box, but also mixed language support for a given query or document. Byte-pair encoding is a proven algorithm for subword tokenization \citep{gage1994new}. More details for the specific approach used in this study is described below.

\subsection{Subword Model}
Following the subword tokenization method used by \citet{fan2021beyond}, a SentencePiece model was trained for 0.9995 character coverage across the hundred most popular character-based languages. Given the sampling probability for each language, $p_l=\frac{D_l}{\sum_{i} D_i}$, monolingual data was added for lower resource languages along with temperature sampling ($T=5$) to rescale the sampling distribution to $p_l^\frac{1}{T}$. The result is a multilingual dictionary with 128k unique tokens representing the given languages. See Appendix~\ref{sec:appendix} for all languages used.

\subsection{English Performance}
Equipped with a multilingual dictionary of fixed size, we are able to tokenize document and query texts into subword units for calculating STF-IDF vectors. Without any whitespace splitting, stop word removal, or word stemming, the STF-IDF approach offers an even higher accuracy of 85.4\% for XQuAD's English passage retrieval as shown in Table~\ref{tab:stfidf}. 

Subword tokens account for word normalization and discounting via the $IDF$ term in a more statistically rigorous machine learning approach than manually crafted stop words and stemming, respectively. To demonstrate how manually crafted rules are no longer necessary, we can apply STF-IDF after such TF-IDF heuristics are applied. The results in Table~\ref{tab:stfidf} can be compared side-by-side with the analogous values in Table~\ref{tab:tfidf}. We find that removing stop words can reduce the signal and hence the accuracy. The effect of stemming prior to subword tokenization has no noticeable effect on accuracy, which reflects how STF-IDF can inherently normalize words. 

\subsection{Multilingual Performance}
To realize the multilingual capabilities of STF-IDF, we performed the same evaluation on the other languages of XQuAD and found that at least 80\% accuracy on 10 other languages is possible without any extra processing steps (see Table~\ref{tab:mstfidf}). We emphasize that these results are based on a single subword model trained on the top 100 character-based languages. 

The performance of each language generally correlates with the subword dictionary coverage for that language, e.g. there is more coverage for Spanish than English, more for English than Chinese, and more for Chinese than Hindi, etc. Nevertheless, these encouraging results demonstrate the promise of STF-IDF for other languages that we do not have evaluation data for.

\subsection{Other Considerations}
Beyond the English performance superiority and broad language coverage, we note that STF-IDF offers other advantages that are beyond the scope of this study:
\begin{itemize}
  \item Subword model can be trained for a specific language or dataset
  \item Subword model can be retrained if data drifts
  \item All vectors are identical in size for a given model
  \item Each vector can encode multiple languages in the default Text2Text model
  \item Vectors can be used for text classification, including language identification
\end{itemize}

\subsection{Text2Text}
We open source our solution and results as part of the Text2Text package available on \href{https://github.com/artitw/text2text}{GitHub} and the \href{https://pypi.org/project/text2text/}{Python Package Index (PyPI)}. A demo Python notebook to reproduce all the results described here is also provided \citep{text2text@2020}.

\section{Conclusion}
While the performance of many TF-IDF search and mining solutions can depend on manually curated recipes, STF-IDF offers a fully machine learnable and extensible approach applicable to languages including English and beyond. One can either train a new subword model specific to an application, or simply use the open-sourced one provided through Text2Text. There remains opportunities to study and quantify the advantages of STF-IDF on documents and queries of more languages and of mixed languages. Additionally, STF-IDF should be evaluated on other information retrieval and mining benchmarks.

\bibliography{anthology,custom}

\begin{thebibliography}{9}
\expandafter\ifx\csname natexlab\endcsname\relax\def\natexlab#1{#1}\fi

\bibitem[{Artetxe et~al.(2019)Artetxe, Ruder, and Yogatama}]{Artetxe:etal:2019}
Mikel Artetxe, Sebastian Ruder, and Dani Yogatama. 2019.
\newblock \href {http://arxiv.org/abs/1910.11856} {On the cross-lingual
  transferability of monolingual representations}.
\newblock \emph{CoRR}, abs/1910.11856.

\bibitem[{Fan et~al.(2021)Fan, Bhosale, Schwenk, Ma, El-Kishky, Goyal, Baines,
  Celebi, Wenzek, Chaudhary et~al.}]{fan2021beyond}
Angela Fan, Shruti Bhosale, Holger Schwenk, Zhiyi Ma, Ahmed El-Kishky,
  Siddharth Goyal, Mandeep Baines, Onur Celebi, Guillaume Wenzek, Vishrav
  Chaudhary, et~al. 2021.
\newblock Beyond english-centric multilingual machine translation.
\newblock \emph{J. Mach. Learn. Res.}, 22(107):1--48.

\bibitem[{Gage(1994)}]{gage1994new}
Philip Gage. 1994.
\newblock A new algorithm for data compression.
\newblock \emph{C Users Journal}, 12(2):23--38.

\bibitem[{Jones(1972)}]{jones1972statistical}
Karen~Sparck Jones. 1972.
\newblock A statistical interpretation of term specificity and its application
  in retrieval.
\newblock \emph{Journal of documentation}.

\bibitem[{Porter(1980)}]{porter1980algorithm}
Martin~F Porter. 1980.
\newblock An algorithm for suffix stripping.
\newblock \emph{Program}.

\bibitem[{Quinonero-Candela et~al.(2008)Quinonero-Candela, Sugiyama,
  Schwaighofer, and Lawrence}]{quinonero2008dataset}
Joaquin Quinonero-Candela, Masashi Sugiyama, Anton Schwaighofer, and Neil~D
  Lawrence. 2008.
\newblock \emph{Dataset shift in machine learning}.
\newblock Mit Press.

\bibitem[{Sennrich et~al.(2015)Sennrich, Haddow, and
  Birch}]{sennrich2015neural}
Rico Sennrich, Barry Haddow, and Alexandra Birch. 2015.
\newblock Neural machine translation of rare words with subword units.
\newblock \emph{arXiv preprint arXiv:1508.07909}.

\bibitem[{Wangperawong(2020)}]{text2text@2020}
Artit Wangperawong. 2020.
\newblock \href {https://github.com/artitw/text2text} {Text2text: Crosslingual
  nlp/g toolkit}.
\newblock \url{https://github.com/artitw/text2text}.

\bibitem[{Widmer and Kubat(1996)}]{widmer1996learning}
Gerhard Widmer and Miroslav Kubat. 1996.
\newblock Learning in the presence of concept drift and hidden contexts.
\newblock \emph{Machine learning}, 23(1):69--101.

\end{thebibliography}
\bibliographystyle{acl_natbib}

\appendix
\section{Languages}
\label{sec:appendix}
Afrikaans (af) \\
Amharic (am) \\
Arabic (ar) \\
Asturian (ast) \\
Azerbaijani (az) \\
Bashkir (ba) \\
Belarusian (be) \\
Bulgarian (bg) \\
Bengali (bn) \\
Breton (br) \\
Bosnian (bs) \\
Catalan Valencian (ca) \\
Cebuano (ceb) \\
Czech (cs) \\
Welsh (cy) \\
Danish (da) \\
German (de) \\
Greeek (el) \\
English (en) \\
Spanish (es) \\
Estonian (et) \\
Persian (fa) \\
Fulah (ff) \\
Finnish (fi) \\
French (fr) \\
Western Frisian (fy) \\
Irish (ga) \\
Gaelic Scottish Gaelic (gd) \\
Galician (gl) \\
Gujarati (gu) \\
Hausa (ha) \\
Hebrew (he) \\
Hindi (hi) \\
Croatian (hr) \\
Haitian Haitian Creole (ht) \\
Hungarian (hu) \\
Armenian (hy) \\
Indonesian (id) \\
Igbo (ig) \\
Iloko (ilo) \\
Icelandic (is) \\
Italian (it) \\
Japanese (ja) \\
Javanese (jv) \\
Georgian (ka) \\
Kazakh (kk) \\
Central Khmer (km) \\
Kannada (kn) \\
Korean (ko) \\
Luxembourgish Letzeburgesch (lb) \\
Ganda (lg) \\
Lingala (ln) \\
Lao (lo) \\
Lithuanian (lt) \\
Latvian (lv) \\
Malagasy (mg) \\
Macedonian (mk) \\
Malayalam (ml) \\
Mongolian (mn) \\
Marathi (mr) \\
Malay (ms) \\
Burmese (my) \\
Nepali (ne) \\
Dutch Flemish (nl) \\
Norwegian (no) \\
Northern Sotho (ns) \\
Occitan (oc) \\
Oriya (or) \\
Panjabi Punjabi (pa) \\
Polish (pl) \\
Pushto Pashto (ps) \\
Portuguese (pt) \\
Romanian Moldavian Moldovan (ro) \\
Russian (ru) \\
Sindhi (sd) \\
Sinhala Sinhalese (si) \\
Slovak (sk) \\
Slovenian (sl) \\
Somali (so) \\
Albanian (sq) \\
Serbian (sr) \\
Swati (ss) \\
Sundanese (su) \\
Swedish (sv) \\
Swahili (sw) \\
Tamil (ta) \\
Thai (th) \\
Tagalog (tl) \\
Tswana (tn) \\
Turkish (tr) \\
Ukrainian (uk) \\
Urdu (ur) \\
Uzbek (uz) \\
Vietnamese (vi) \\
Wolof (wo) \\
Xhosa (xh) \\
Yiddish (yi) \\
Yoruba (yo) \\
Chinese (zh) \\
Zulu (zu) \\

\end{document}